\documentclass[final,12pt]{elsarticle}

\usepackage{amssymb}

 \usepackage{booktabs,makecell}  
\usepackage{amsmath,graphicx}
\usepackage{threeparttable}
\usepackage{booktabs}
\usepackage{amsmath,graphicx}
\usepackage{url}
\usepackage{psfrag}
\usepackage{amssymb}
\usepackage{amsbsy}
\usepackage{graphicx}
\usepackage{array}
\usepackage{multirow}
\usepackage{bm}
\usepackage{subcaption}
\usepackage{verbatim}
\usepackage{relsize}
\usepackage{algorithm}
\usepackage{algpseudocode}
\usepackage{tabularx}
 \usepackage{verbatim}
\usepackage{bm}
\usepackage{lipsum}
\usepackage{tabularx}
\usepackage{mwe} 
\usepackage[utf8]{inputenc} 
\usepackage[T1]{fontenc}    
\usepackage{hyperref}       
\usepackage{url}            
\usepackage{booktabs}       
\usepackage{nicefrac}       
\usepackage{microtype}      
\usepackage{times}
\usepackage{epsfig}
\usepackage{graphicx}
\usepackage{amsmath}
\usepackage{amsthm}
\usepackage{mathrsfs} 
\usepackage{amssymb}
   \usepackage[T1]{fontenc}
    \usepackage{mathpazo}

    \usepackage{graphicx}

\theoremstyle{plain}

\theoremstyle{definition}

\def\x{\mathbf{x}}

\def\a{\mathbf{a}}

\def\v{\mathbf{v}}

\def\p{\mathbf{p}}

\def\z{\mathbf{z}}

\def\R{\mathbb{R}}

\def\balpha{\boldsymbol{\alpha}}

\def\x{{\mathbf x}}

\usepackage{color}

\usepackage{caption}
\usepackage{wrapfig}
\usepackage{tabularx}
\usepackage{placeins}
\usepackage{dblfloatfix}
\usepackage[table]{xcolor}

\definecolor{mycolor}{cmyk}{0.0,0.2,0.9,0.0}

\journal{Machine Learning with Applications}

\begin{document}

\begin{frontmatter}



\title{Zero-Shot Image Classification Using Coupled Dictionary Embedding}


\author{Mohammad Rostami$^{1,*}$, Soheil Kolouri$^{2,*}$\footnote{Equal Contribution}, Zak Murez$^3$, Yuri Owechko$^4$,  Eric Eaton$^5$, Kuyngnam Kim$^6$\\
mrostami@isi.edu;soheil.kolouri@vanderbilt.edu;zak@murez.com\\yowechko@hrl.com,eeaton@seas.upenn.edu;ai.guru@sk.com}

\address{$^1$USC Information Sciences Institute, CA, USA\\$^2$Vanderbilt University, Nashville, TN, USA\\$^3$  Wayve, London, UK\\ HRL Laboratories, Malibu,  CA, USA,\\ $^5$ University of Pennsylvania, Philadelphia, PA, USA
\\$^6$ SK Telecom, Seoul, South Korea\\ \clearpage}

\begin{abstract}
Zero-shot learning (ZSL)   is a framework to classify images that belong to unseen visual classes using their  semantic descriptions about the unseen classes. 
We develop a new ZSL algorithm   based on coupled dictionary learning. The core idea is to enforce the visual features and the semantic attributes of an image to share the same sparse representation in an intermediate embedding space, modeled as the shared input space of two sparsifying dictionaries. In the ZSL training stage, we  use images from a number of seen classes for which we have access to both the visual and the semantic attributes  to train two coupled dictionaries that  can represent  both the visual and the semantic feature vectors of an image using a single sparse vector. 
In the ZSL testing stage and in the absence of labeled data,  images from   unseen classes are mapped into the attribute space by finding the joint-sparse representations using solely the visual dictionary via solving a LASSO problem. The image is then classified in the attribute space given semantic descriptions of unseen classes. We also provide  attribute-aware and transductive formulations  to tackle the ``domain-shift'' and the ``hubness'' challenges for ZSL, respectively. Experiments on four primary datasets using VGG19 and GoogleNet visual features, are provided. Our performances using VGG19 features are 91.0\%, 48.4\%, and 89.3\% on the SUN , the CUB, and the AwA1 datasets, respectively. Our performances on the SUN, the CUB, and the AwA2 datasets are 57.0\%,49.7\%, and 71.7\%, respectively, when GoogleNet features are used. Comparison with existing methods  demonstrates that our method is effective and compares favorably against the sate-of-the-art. In particular, our algorithm leads to decent performance on the all four datasets. \footnote{Early partial results of this paper is presented at 2018 AAAI~\cite{kolouri2018joint}}.

\end{abstract}


\end{frontmatter}




 \section{Introduction}
 
 Advances in deep learning have led to a remarkable performance improvement in a wide range of  classification and categorization tasks. 
 This success is primarily due to the fact that deep neural networks automate the process of feature engineering using  a blind end-to-end supervised training procedure~\cite{morgenstern2014properties,lecun2015deep}. However, the cost for this success is the need for huge annotated datasets to implement supervised training. Emergence of crowdsourcing data annotation platforms such as Amazon Mechanical Turk has made data annotation easier~\cite{rostami2018crowdsourcing}, but manual data annotation is still not  feasible in many cases, including:
 \begin{enumerate}
   \item  Data annotation for tine-grained multi-class classification (e.g., thousands of classes for animal categorization) is a challenging task because it requires training annotators to be able to provide accurate labels. Additionally, the process  is more time-consuming and usually requires more annotators because the labels are noisier. 
   \item In many domains, including medical domains, sharing data with annotators is not easy due to privacy regulations that limits sharing data. As a result, it is highly challenging to hire annotators that have sufficient clearances to process data.
   \item When the domain is a specialized domain, e.g., synthetic aperture radar images, only people with years of prior training are able to annotate data. As a result, qualified data annotators are limited and expensive to hire. 
    \item The persistent and dynamic emergence of new classes (e.g., new products on shopping websites) makes data annotation a continual time-consuming procedure. Additionally, retraining the model to incorporate new classes can be computationally expensive.
    \item  In some application, there are classes with highly infrequent  members (e.g., rare event classification). Preparing training instances for rare event annotation is challenging.
\end{enumerate}
 
Consequently, training deep learning model is not a feasible solution when the above and  other potentially similar challenges are present. To circumvent these challenges, it is  desirable to improve existing systems by enabling them to benefit from knowledge transfer. For example, it is desirable to learn using a few training samples~\cite{snell2017prototypical,sung2018learning,sun2019meta} and even learning {\em unseen} classes with no accessible training samples~\cite{palatucci2009zero,socher2013zero,norouzi2013zero,lampertattribute,zhang2015zero,long2018zeroo,lu2018attribute}. Additionally, learning from past experiences accumulatively, i.e., continual or lifelong learning, can help to avoid learning redundant information~\cite{chen2018lifelong,zenke2017continual,rostami2019complementary,rostami2020generative,lomonaco2021avalanche,rostami2018multi}.  Such abilities help to 
 classify new emerging classes more efficiently, to relax the need for persistent data annotation and model retraining, and to benefit from past learned experiences~\cite{rostami2021detection}.

\begin{figure} 
\centering
\begin{subfigure}[b]{\textwidth}
        \includegraphics[width=.95\columnwidth]{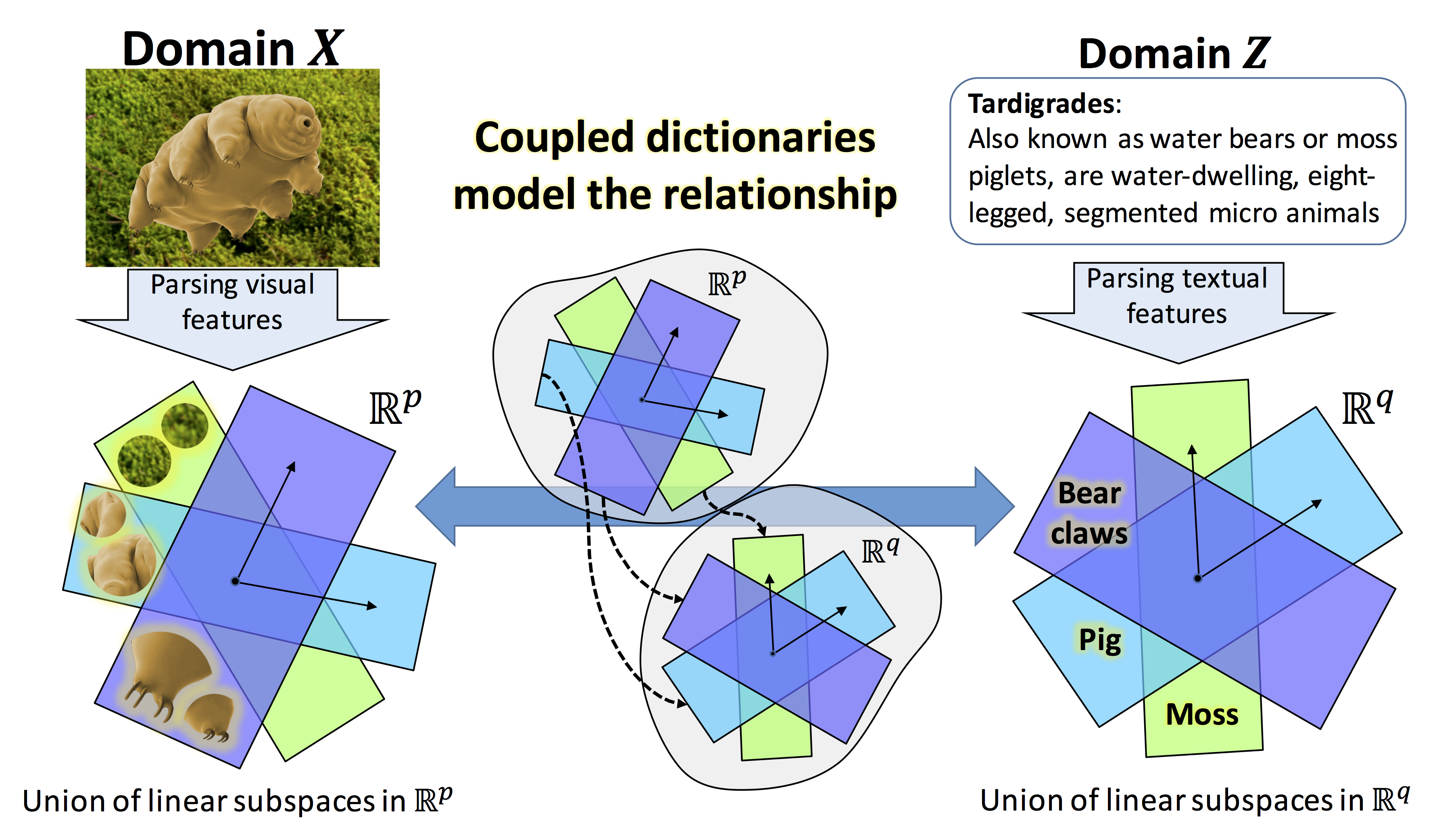}   
          \centering
        \caption{Training}
        \label{ICMLDALfig:1}
        \end{subfigure}
        \begin{subfigure}[b]{\textwidth}
           \includegraphics[width=.95\columnwidth]{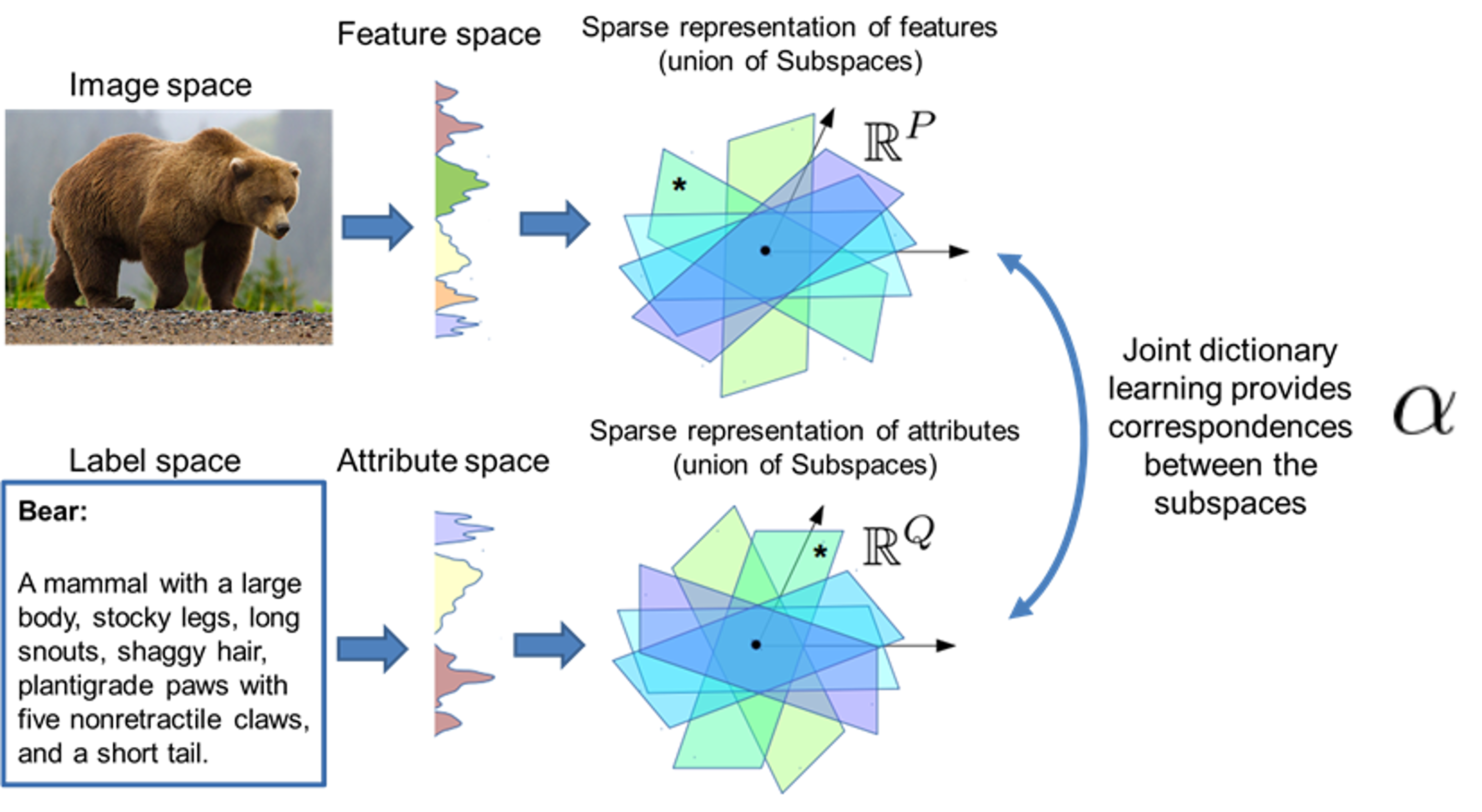}  
             \centering
        \caption{Testing}
        \label{ICMLDALfig:MNISTUSPS}
           \end{subfigure}
   \caption{High-level overview of our  approach: (a) we consider that the visual features (left) and the attribute features (right) can be represented sparsely  using  unions of subspaces that are modeled using two dictionaries. We train the two dictionaries such that these two space are matched (middle), leading to coupling the visual and the attribute features in the shared space. (b) During testing, we solve for the sparse representation of the input images and then find its corresponding semantic description, y first solving for the joint sparse vector using the visual dictionary and then searching for the closest semantic description.}
\label{fig:tardigrade}
\end{figure}

A primary approach to reduce dependence on large annotated datasets  is to benefit from secondary domains of information~\cite{rostami2021transfer}.
Domain adaptation~\cite{pan2011domain,pinheiro2018unsupervised,saito2018maximum,rostami2020sequential,stan2021unsupervised} and zero-shot learning (ZSL)~\cite{palatucci2009zero,socher2013zero,norouzi2013zero,lampertattribute,zhang2015zero,long2018zeroo,lu2018attribute} are two primary learning settings to benefit from auxiliary domains to relax data annotation. Most works within domain adaptation use the strong assumption that the two domains are homogeneous and also share the same set of classes. It is assumed that we have fully annotated data in one of the domains and in the second domain only unannotated data is accessible. The goal would be to train a classifier for the unannotated domain by traninferring knowledge from the annotated domain~\cite{rostami2019sar,rostami2021CL}. In contrast, zero-shot learning considers learning unseen classes in a single domain, usually visual data, by coupling it with an auxiliary domain, usually natural language information, using a number of seen classes for which we have bi-view annotations for both domains. We focus on ZSL in this work. ZSL is inspired by the ability of humans to recognize new visual classes using their semantic descriptions.
Humans remarkably are extremely good at learning enormous numbers of classes from little data using descriptions in natural language. Consider the problem of classifying animal images. It is estimated that as many as one million different species of animals have been identified, with as many as ten thousand new species being discovered annually. This classification problem is a case when ZSL can be extremely helpful. Most people probably have not seen an image of a `tardigrade', nor heard of this species. If you have not heard of this animal specie, we can intuitively demonstrate how ZSL can be possible for this class.
Consider the following sentence from Wikipedia: ``Tardigrades''  (also known as water bears or moss piglets) are water-dwelling, eight-legged, segmented micro animals.'' Given this textual description, most humans can easily identify the creature in Figure~\ref{fig:tardigrade} (a) (left) as a Tardigrade, even though they may have never seen one before. Humans can easily perform this ZSL task by: 1) identifying the semantic features that describe the class {\it Tardigrade} as `bear-like', `piglet-like', `water-dwelling', `eight-legged', `segmented', and `microscopic animal', 2) parsing the image into its visual attributes (see Figure~\ref{fig:tardigrade} (a)), and 3) matching the parsed visual features to the parsed textual information. In other words, humans can transfer knowledge from the domain of natural language to solve problems in the vision domain.

 ZSL has been implemented in computer vision based on the above intuition. To this end, we can parse textual features into a vector of either predetermined binary attributes (e.g., water-dwelling) or   continues features using, e.g., using {\it word2vec}~\cite{mikolov2013distributed}. We can also use pretrained deep convolutional neural networks (CNNs)   to extract   visually rich   features from natural images to parse the visual information.  ZSL algorithms generally learn a mapping between the visual features and semantic attributes using a shared intermediate embedding space~\cite{palatucci2009zero,socher2013zero,norouzi2013zero,lampertattribute,zhang2015zero,long2018zeroo,lu2018attribute}.  After model training, his intermediate space can be used to transfer knowledge across the two domains. ZSL has been found to be practically beneficial on applications, including face verification~\cite{kumar2009attribute}, video
understanding~\cite{wu2016harnessing}, and object recognition~\cite{lampert2013attribute}.
We can categorize the ZSL algorithms into two primary subgroups. A group of ZSL methods model the cross-domain mapping as a linear function~\cite{frome2013devise,romera2015embarrassingly,akata2013label,akata2015evaluation}. Since a simple hypothesis space is used to learn the cross-domain mapping, learning the cross-domain mapping is computationally efficient. However, nonlinear relations between the domains may not be encoded well. In contrast, more recent methods use deep neural networks to model the mapping~\cite{xian2016latent,xian2018feature,liu2018generalized,li2019leveraging}. 
Although deep neural networks usually lead to state-of-the-art performance for diverse set of applications, training a neural network for ZSL will require more data instances of seen classes which goes against the very goal of ZSL. ZSL methods that can learn the cross-domain mapping as a nonlinear function and at the same time do not have significant complexity are desirable.
In this paper, we follow a middle-ground between the above two subgroups to develop a ZSL algorithm which has a competitive performance despite having a less computational model training complexity.

We develop a new ZSL algorithm based on coupled dictionary learning (CDL)~\cite{yang2010image} to relate the visual features and the semantic attributes.
CDL in essence is a (semi-)linear model but it can encode nonlinearities. This ability stems from the fact that the hypothesis space of a dictionary is the union of linear subspaces, i.e., a nonlinear   space.
Our specific contributions include:
\begin{enumerate}
    \item We formulate ZSL as a coupled dictionary learning problem and demonstrate by solving a dictionary learning problem, ZSL can be performed.
    \item We provide an efficient algorithm to solve the resulting joint dictionary learning optimization problem.
    \item We also address the challenges of hubness~\cite{dinu2014improving} and domain-shift~~\cite{fu2015transductive} in ZSL using by augmenting our base optimization problem with suitable regularization terms.
    \item We provide theoretical analysis which establishes PAC-learnability of our proposed algorithm.
    \item We perform experiments on primary benchmark datasets and demonstrate that our method is effective and compares favorably with respect to the state-of-the-art.
\end{enumerate}

 The remaining of the paper is as follows.
 In section~2, we will explain the formulation we use for ZSL and our high-level algorithmic idea. Section~3 summarizes our algorithm to tackle ZSL. Section~4 provides a theoretical analysis for our algorithm. We have provides experimental validation in section~5. The paper finally is concluded in section~6 with a brief discussion.

\section{Problem Formulation and Technical Rationale}

We follow Palatucci et. al. to formulate ZSL as a two stage estimation problem~\cite{palatucci2009zero}. Consider a visual feature metric space $\mathcal{F}$ of dimension  $p$, a semantic metric space $\mathcal{A}$ with dimension of $q$ as well as a class label set $\mathcal{Y}$ with dimension $K$ which ranges over a finite alphabet of size $K$ (images can potentially have multiple memberships in the classes). As an example $\mathcal{F}=\mathbb{R}^p$ for the visual features extracted from a deep CNN and $\mathcal{A}=\{0,1\}^q$ when a binary code of length $q$ is used to identify the presence/absence of various characteristics in an object~\cite{lampertattribute}. We are given a labeled dataset $\mathcal{D}=\{((\bf{x}_i,\bf{z}_i),\bf{y}_i)\}_{i=1}^N$ of features of seen images and their corresponding semantic attributes, where $\forall i:\bf{x}_i\in \mathcal{F}, \bf{z}_i\in \mathcal{A}$, and $\bf{y}_{i}\in \mathcal{Y}$.  We are also given the unlabeled attributes of unseen classes  $\mathcal{D}'=\{\bf{z}'_j ,\bf{y}'_j\}_{j=1}^M$ (i.e., we have access to textual information for a wide variety of objects but not have access to the corresponding visual information).  Following the standard assumption in ZSL, we   assume that the set of the seen and the unseen classes are disjoint. The challenge is how to learn a model on the labeled set  and transfer the learned knowledge to the unlabeled set.  We also assume that the same semantic attributes could not describe two different classes of objects, i.e., by knowing semantic attribute of an image one can classify that image. The goal is to learn from the labeled dataset   to classify images of unseen classes.  For further clarification, consider an instance of ZSL in which features extracted from images of horses and tigers are included in seen visual features $X=[\bf{x}_1,...,\bf{x}_N]$, where $\bf{x}_i\in\mathcal{F}$, but $X$ does not contain features of zebra images. On the other hand, the semantic attributes contain information of all the seen images $Z=[\bf{z}_1,...,\bf{z}_N]$ for $\bf{z}_i\in\mathcal{A}$ and the unseen images $Z'=[\bf{z}'_1,...,\bf{z}'_M]$ for $\bf{z}'_j\in\mathcal{A}$   including the zebras. The goal is   by learning the relationship  between the image features and the attributes ``horse-like'' and ``has stripes'' from the seen images, we   assign an unseen zebra image to its corresponding attribute.

Within this paradigm, ZSL can be performed by a two stage estimation. First   the visual features can be mapped into the semantic space and then the label is estimated in the semantic space.   More formally,  we want to learn the  mapping  $\phi:\mathcal{F}\rightarrow\mathcal{A} $ which relates the visual space and the attribute space. We also assume that $\psi:\mathcal{A}\rightarrow \mathcal{Y}$ is the mapping between the semantic space and the label space. The mapping $\psi$ can be as simple as nearest neighbor, i.e., we assign labels according to the closest semantic attribute in the semantic attribute space.  Having learned this mapping, one can recover the corresponding attribute vector for an unseen image using the image features and then classify the image using a second mapping $\bm{y}=(\psi\circ  \phi)(\bf{x})$, where `$\circ$' represents function composition. The goal is to introduce a type of bias to learn both mappings using the labeled dataset. Having learned both mappings,  ZSL is   feasible in the testing stage. Because if the mapping $\phi(\cdot)$ can map an unseen image close enough to its true semantic features, then intuitively the mapping $\psi(\cdot)$ can still recover the corresponding class label. 
 Following our example, if the function $\phi(\cdot)$ can recover that an unseen image of a zebra is ``horse-like'' and ``has stripes'', then it is likely that the mapping $\psi(\cdot)$ can classify the unseen image. Our core idea is to benefit from coupled dictionary learning~\cite{yang2010image} to model these mappings.

\subsection{Proposed Idea} 
The idea of using coupled dictionaries to map data from a given metric space to a second related metric  space was first proposed by Yang et al. ~\cite{yang2010image} for single image super-resolution problem~\cite{rehman2012ssim}. Their pioneer idea is to assume that the high-resolution and the corresponding low-resolution patches of   image  can be represented with a unique joint sparse vector in two low- and high-resolution dictionaries. The core idea is that in the absence of a high-resolution image and given its low-resolution version, the corresponding joint sparse representation can be computed using sparse signal recovery. 
The sparse vector is  then can be used to generate the high-resolution image patches using the low-resolution image. They also propose an efficient algorithm to learn the low- and the high-resolution dictionaries using a training set, consisting of both low- and high-resolution version of natural images. Our goal is to follow the same approach but replacing the low- and the high-resolution metric spaces with the visual and the semantic spaces, respectively.

As a big picture to understand our approach, 
Figure \ref{fig:tardigrade} captures the gist of our idea. In Figure \ref{fig:tardigrade} (a), visual features are extracted via CNNs (left sub-figure). For example, the last fully-connected layer of a trained CNN can be removed and the rest of the deep net can be used as a feature extractor given an input image. These features have been demonstrated to be highly descriptive and lead to the state-of-the-art performance for many computer vision and image processing tasks. To perform ZSL, we need textual description of the classes, too. Textual description of many classes are cheap to obtain, e.g., Wikipedia.
The semantic attributes then can be provided via textual feature extractors like word2vec or potentially via human annotations (right sub-figure). It is assumed that 
both the visual features and the semantic attributes can be represented   sparsely  using the visual and the attribute dictionaries that is modeled  as a shared union of linear subspaces   (left and right sub-figures). The  idea here is that the sparse representation vectors for both feature vectors are equal. Thus, one can map an image to its textual description  in this space using the joint sparse vector (middle sub-figure). 

The intuition from  a co-view  perspective  is that both the visual and the attribute features provide information about the same class or entity, and so each can augment  learning of the other to improve performance~\cite{isele2016using,rostami2019using}. Each underlying class is common to both views, and so we can find task embeddings that are consistent for both the visual features  and their corresponding attribute features. The main  challenge is to learn these dictionaries for the visual and the   attribute spaces. Having learned these two dictionaries,   zero-shot classification can be performed by mapping images of unseen classes into the attribute space, where classification can be simply done via nearest neighbor or more advanced clustering approaches (Figure \ref{fig:tardigrade} (b)). Given the coupled nature of the learned dictionaries, an image can be mapped to its semantic attributes by first finding the sparse representation with respect to the visual dictionary. Our algorithm is equipped with a novel entropy minimization regularizer~\cite{guo2006use}, which facilitates a better solution for  the ZSL problem. The entropy regularization helps to tackle the challenge of domain-shift in zero-shot learning. Next  the semantic attribute dictionary can be used to recover the attribute  vector from the joint sparse representation which can then be used for classification (Figure \ref{fig:tardigrade} (b)).   We also show that a transductive approach applied to our attribute-aware JD-ZSL formulation improves the perofrmance via mitigating the challenge of hubness in high dimensions. Our experiments demonstrate that our algorithm leads to competitive performance on four standard ZSL benchmark datasets.

\subsection{Technical Rationale}

 For the rest of our discussion we assume that $\mathcal{F}=\R^p$, $\mathcal{A}=\R^q$, and $\mathcal{Y}\subset \R^K$. Most ZSL algorithms focus on learning $\phi(\cdot)$ because even using a simple method like nearest neighbor classification for $\psi(\cdot)$  yields descent ZSL performance.  The simplest ZSL approach is to assume that the mapping $\phi:\mathbb{R}^p\rightarrow \mathbb{R}^q$ is linear, $\phi(\x)=W^T\x$ where $W\in \R^{p\times q}$, and then minimize the regression error $\frac{1}{N}\sum_i \|W^T\x_i-\z_i\|^2_2$ to learn $W$. Although closed form solution exists for $W$, the solution contains the inverse of the covariance matrix of $X$, $(\frac{1}{N}\sum_i (x_ix_i^T))^{-1}$, which requires a large number of data points for accurate estimation. To overcome this problem, various regularizations are considered for $W$. Decomposition of $W$ as $W=P\Lambda Q$, where $P\in\R^{p\times l}$, $\Lambda\in\R^{l\times l}$, $Q\in\R^{l\times q}$, and $l<min(p,q)$ can   be helpful. Intuitively, $P$ is a right linear operator that projects $\x$'s into a shared low-dimensional subspace, $Q$ is a left linear operator that projects  $\z$ into the same shared subspace, and $\Lambda$ provides a bi-linear similarity measure in the shared subspace. The  regression problem can then  be transformed into maximizing $\frac{1}{N}\sum_i \x_i^TP\Lambda Q\z_i$, which is a weighted correlation between the embedded $\x$'s and $\z$'s. This is the essence of many ZSL techniques including   Romera-Paredes et al.~\cite{romera2015embarrassingly}.  This technique  can be   extended to nonlinear mappings using  kernel methods. However, the choice of kernels   remains  an open challenge.

 The mapping $\phi:\R^p\rightarrow \R^q$ can also be chosen to be highly nonlinear, e.g.,  deep neural networks.   Let a deep net be denoted by $\phi(.;\bm{\theta})$, where $\bm{\theta}$ represents the synaptic weights and biases. ZSL can then be addressed by minimizing $\frac{1}{N}\sum_i \|\phi(\x_i;\bm{\theta})-\z_i\|^2_2$ with respect to $\bm{\theta}$. Alternatively, one can nonlinearly embed $\x$'s and $\z$'s in a shared metric space via deep nets, $p(\x;\bm{\theta}_x):\R^p\rightarrow \R^l$ and $q(\z;\theta_z):\R^q\rightarrow \R^l$, and maximize their similarity measure in the embeding space, $\frac{1}{N}\sum_i p(\x_i;\bm{\theta}_x)^T q(\z_i;\bm{\theta}_z)$.
 This approach might improve performance for particular data sets, but in turn would require more training samples. Note however, this might not be plausible for ZSL because the very reason and motivation to perform ZSL is to learn from as few labeled data points as possible.   

Comparing the above approaches, nonlinear methods   are computationally expensive and require a large   training dataset. 
In contrast, linear ZSL algorithms are  efficient, but their  performances are   lower.  As a compromise, we can model nonlinearities in data distributions as a union of linear subspaces using coupled dictionaries~\cite{das2019zero}.
The relationship between the metric spaces is also reflected in the learned dictionaries. This allows a nonlinear scheme with a computational complexity comparable to   linear techniques.

\section{Zero-Shot Learning using Coupled Dictionary Learning}
\label{sec:jointDL}

In the standard dictionary learning framework, a sparsifying dictionary is learned using a given training sample set $X=[\x_1,...,\x_N]$ for a particular class of signals. 
Unlike standard dictionary learning, coupled dictionary learning has been proposed to couple related features from two   metric spaces to learn the mapping function between these spaces.  
Following the same framework, the gist of our approach is to learn the  mapping  $\phi:\mathbb{R}^p\rightarrow \mathbb{R}^q$ through two dictionaries, $D_x\in\mathbb{R}^{p\times r}$ and $D_z\in \mathbb{R}^{q\times r}$  for $X$ and $[Z,Z']$ sets, respectively, where $r>max(p,q)$. The goal is to find a shared sparse representation $\a_i$ for $\x_i$ and $\z_i$, such that $\x_i=D_x\a_i$ and $\z_i=D_z\a_i$. The shared sparse representation  couples the semantic and visual feature spaces.  We first explain the training procedure for the two dictionaries, and then the way we use these dictionaries to estimate $\phi(\cdot)$.

\subsection{Training Phase}

The standard dictionary learning is based upon minimizing the empirical average estimation error $\frac{1}{N}\|X-D_xA\|^2_F$ on a given training set $X$, where an additive $\ell_1$ regularization penalty term on $A$ enforces sparsity:
\begin{equation}
\begin{split}
 D^*_x,A^* = &  \operatorname*{argmin}_{D_x,A} \left\{\frac{1}{N}\|X-D_xA\|^2_F+\lambda\|A\|_1 \right\} \\&
  \hspace{.2in}\text{s.t.}~ \bigl\|D^{[i]}_x\bigr\|^2_2 \leq 1 \enspace.
\label{eq:mainDx}
\end{split}
\end{equation}  
 Here $\lambda$ is the regularization parameter that controls sparsity level and $D_x^{[i]}$ is the $i^{th}$ column of $D_x$. The columns of the dictionary are normalized to obtain a unique dictionary.  Alternatively, following the Lagrange multiplier technique, the Frobenius norm of $D_x$ could be used as a regularizer in place of the constraint.
 The above problem is not a convex optimization problem, but is convex in each variable alone; it is biconvex with respect to the variables $D_x$ and $A$. As a result, most dictionary learning algorithms use alternation on variables $D_x$ and $A$ to solve~\eqref{eq:mainDx} which leads to iterations on two separate optimizations. Each optimization problem is performed solely on one of the variables, assuming the other variable to be constant. Upon a suitable initialization, Eq.~\eqref{eq:mainDx} reduces to a number of parallel sparse recovery when the dictionary is fixed, i.e., LASSO problems which can be solved efficiently. Then, for a fixed $A$, Eq.~\eqref{eq:mainDx} reduces   to a standard quadratically constrained quadratic program (QCQP) problem which can be solved efficiently with iterative methods such as  conjugate gradient descent algorithms   even for high-dimensional (large $p$) and huge problems (large $r$). This alternative procedure on the variables is repeated until some convergence criterion is met.

 In our coupled dictionary learning framework, we aim to learn coupled dictionaries $D_x$ and $D_z$ such that they share the sparse coefficients $A$ to represent the seen visual features $X$ and their corresponding attributes $Z$, respectively. Intuitively this means that visual features for an object  have corresponding semantic features. On the other hand, $D_z$ also needs to sparsify the semantic attributes of other (unseen) classes, $Z'$, to enable performing ZSL. Hence, we propose the following   optimization problem to learn both dictionaries:
 \small
\begin{equation}
\begin{split}
&D^*_x,A^*,D^*_z,B^*  =   \underset{D_x,A,D_z,B}{\operatorname*{argmin}}  \biggl\{ \frac{1}{Np}\biggl(\|X-D_xA\|^2_F+ \frac{p\lambda}{r}\|A\|_1\biggr)   \\& +\frac{1}{Nq}\|Z-D_zA\|^2_F+ 
  \frac{1}{Mq}\biggl(\|Z'-D_zB\|^2_F+\frac{q\lambda}{r}\|B\|_1\biggr) \biggr\} \\&
  \hspace{.2in} \text{s.t.:}  \bigl\|D^{[i]}_x\bigr\|^2_2\leq 1,~\bigl\|D^{[i]}_z\bigr\|^2_2\leq 1 \enspace.
\label{eq:maineq}
\end{split}
\end{equation}  
\normalsize 
The above formulation combines the dictionary learning problems for $X$ and $Z$ by coupling them via the joint sparse code matrix $A$, and also enforces $D_z$ to be a sparsifying dictionary for $Z'$ with the sparse codes $B$. Similar to Eq.~\eqref{eq:mainDx}, the optimization in Eq.~\eqref{eq:maineq}
 is biconvex in $(D_x,D_z)$ and $(A,B)$. Hence, we use an alternative scheme to update  $D_x$ and $D_z$ for a local solution. 

 First we add the constraints on dictionary atoms in~\eqref{eq:maineq} as a penalty term:
\begin{equation}
\min_{A,D_x}||X-D_xA||_2^2+\lambda ||A||_1+\beta ||D_x||_2^2
\label{eq:opt1}
\end{equation}
We can solve Eq.~\eqref{eq:opt1} by alternately solving   LASSO for $A$ and taking gradient steps with respect to $D_x$. 
Next we solve the following problem:
\begin{equation}
\min_{B,D_z}||Z-D_zA||_2^2+||Z'-D_zB||_2^2+\lambda ||B||_1+\beta ||D_z||_2^2
\label{eq:opt2}
\end{equation}
by alternately solving the LASSO for $B$ and taking gradient steps with respect to $D_z$, (while holding $A$ fixed as the solution found in Eq.~\eqref{eq:opt1}. 

Algorithm~\ref{algo1} summarizes the coupled dictionary learning procedure. The learned dictionaries then can be used to perform ZSL in the testing phase (see Figure~1 (b)). 

\begin{algorithm}[tb!]
\caption{\ Coupled Dictionary Learning ($\{X,Z,Z'\}$, $\lambda$, $r$,$itr$)}
\label{algo1}
\begin{algorithmic}[1]
\State $D_x \gets  \bm{\text{RandomMatrix}}_{p,r}, \ D_z \gets  \bm{\text{RandomMatrix}}_{q,r}$
\State $D_x \gets$ update$(D_x,  \{X\} , \lambda, \beta)$ \hspace{82pt} Eq.~\ref{eq:opt1}
\State $D_z \gets$ update$(D_z,  \{ Z, Z', A \} , \lambda, \beta)$ \hspace{58pt} Eq.~\ref{eq:opt2}
\end{algorithmic}
\end{algorithm}

\subsection{Prediction of Unseen Attributes}

In the testing phase, we are only given the extracted features from unseen images  and the goal is to predict their corresponding semantic attributes. We propose two different methods to predict the semantic attributes of the unseen images based on the learned dictionaries in the training phase, namely attribute-agnostic  prediction and attribute-aware prediction methods. 

\subsubsection{Attribute-Agnostic Prediction}

The attribute-agnostic (AAg) method  is the naive way of predicting the semantic attributes from an unseen image $\x'_i$. In the attribute-agnostic formulation, we first find the sparse representation $\balpha_i$ of the unseen image $\x'_i$ by solving the following LASSO problem, 
\begin{equation}
\balpha_i=\operatorname{argmin}_\a\biggl\{\frac{1}{p}\|\x_i-D_x\a\|_2^2 +\frac{\lambda}{r}\|\a\|_1\biggr\}\enspace.
\label{eq:attrAgn}
\end{equation}
and its corresponding attribute is estimated by $\hat{z}_i=D_z\balpha_i$. We call this formulation attribute-agnostic because  the sparse coefficients are found without any information from the attribute space. We use AAg as a baseline to demonstrate the effectiveness of   the attribute-aware prediction.

\subsubsection{Attribute-Aware Prediction}

In the attribute-aware (AAw) formulation, we would like to find the sparse representation $\balpha_i$ to not only approximate the input visual feature, $\x'_i\approx D_x\balpha_i$, but also provide an attribute prediction, $\hat{z}_i=D_z\balpha_i$, that is well resolved in the attribute space. This mean that   ideally we want to have $\hat{\z}_i=\z'_m$, for some $m\in\{1,...,M\}$. To achieve this, we define the soft assignment of $\hat{\z}_i$ to $\z'_m$, denoted by $p_m$, using the Student's t-distribution as a kernel to measure similarity between $\hat{\z}_i=D_z\balpha_i$ and $\z'_m$,
\begin{equation}
p_m(\balpha_i)=\frac{\biggl(1+\frac{\|D_z\balpha_i-\z'_m\|^2_2}{\rho}\biggr)^{-\frac{\rho+1}{2}}}{\mathlarger{\mathlarger{\sum}}_k \biggl(1+\frac{\|D_z\balpha_i-\z'_k\|^2_2}{\rho}\biggr)^{-\frac{\rho+1}{2}}}\enspace,
\label{eq:softass} 
\end{equation}
where $\rho$ is the kernel parameter. We chose t-distribution as it is less sensitive to the choice of kernel parameter, $\rho$. 

Ideally, $p_m(\balpha_i)=1$ for some $m\in\{1,...,M\}$ and $p_j(\balpha_i)=0$ for $j\neq m$. In other words, the ideal soft-assignment $\p=[p_1,p_2,...,p_M]$ would be one-sparse and hence would have minimum entropy. This motivates our attribute-aware formulation, which penalizes Eq.~\ref{eq:attrAgn} with the entropy of $\p$.
\begin{equation}
\begin{split}
\balpha_i= \operatorname{argmin}_\a \biggl\{\underbrace{\frac{1}{p}\|\x'_i-D_x\a\|_2^2 -\gamma \sum_m p_m(\a)\log(p_m(\a))}_{g(\a)}  +\frac{\lambda}{r}\|\a\|_1\biggr\}\enspace,
\end{split}
\label{eq:attrAware}
\end{equation}
where $\gamma$ is the regularization parameter for entropy of the soft-assignment probability vector $\p$, $H_{\bm{p}}(\balpha) $. The entropy minimization has been successfully used in several works~\cite{guo2006use} either as a sparsifying regularization or to boost the confidence of classifiers. Such regularization, however, turns the optimization in Eq.~\eqref{eq:attrAware} into a nonconvex problem.
 However, since $g(\a)$ is differentiable and the $\ell_1$ norm is continuous,   we can apply proximal gradient descent~\cite{parikh2014proximal} or ADMM~\cite{boyd2011distributed} to solve it. 
 In practice, we found that  gradient descent applied directly on Eq.~\eqref{eq:attrAware} works fine since Eq.~\eqref{eq:attrAware} is differentiable almost everywhere. 
Note, however, a good initialization is needed for achieving an accurate solution, due to the non-convex nature of the objective function. Therefore we initialize $\balpha$ from the solution of the attribute-agnostic formulation which is an approximate solution. Finally the corresponding attributes are estimated as $\hat{z}_i=D_z\balpha_i$, for $i=1,...,l$.

 \begin{algorithm}[tb!]
\caption{\ Zero-shot Prediction ($\x_i$ $\lambda$)}
\label{algo2}
\begin{algorithmic}[1]
\State Attribute-Agnostic prediction:
\State $\balpha_i \gets \operatorname{argmin}_\a \frac{1}{p}\|\x_i-D_x\a\|_2^2 +\frac{\lambda}{r}\|\a\|_1$
\State $\hat{\z}_i=D_z\balpha_i$
\State $\z'_m= \operatorname{argmin}_{\z'\in Z'} \|\z'-\hat{\z}_i\|_2$
\State Attribute-Aware prediction:
\State $\balpha_i \gets \operatorname{argmin}_\a  \frac{1}{p}\|\x'_i-D_x\a\|_2^2 -\gamma H_{\bm{p}}(\balpha)  +\frac{\lambda}{r}\|\a\|_1$
\State $\hat{\z}_i=D_z\balpha_i$
\State $\z'_m= \operatorname{argmin}_{\z'\in Z'} \|\z'-\hat{\z}_i\|_2$
\State Transducer Prediction
\State Solve \ref{eq:attrAware} to predict $\balpha_i$ for all unseen samples.
\State Use label propagation to spread  the labels.
\end{algorithmic}
\end{algorithm}

\subsection{From Predicted Attributes to Labels}

To predict the image labels, one needs to assign the predicted attributes  to the $M$ attributes of the unseen classes $Z'$. We performed this task in two approaches, namely the inductive approach and the transductive approach. 
\subsubsection{Inductive Approach}

In the inductive approach, the inference can be performed using a nearest neighbor (NN) approach in which the label of each individual $\hat{\z}_i$ is assigned to be the label of its nearest neighbor $\z'_m$:
\begin{equation}
\z'_m= \operatorname{argmin}_{\z'\in Z'}\biggl\{  \|\z'-\hat{\z}_i\|_2\biggr\}\enspace,
\label{nearest}
\end{equation}
and the corresponding label of $\z'_m$ is assigned to $\hat{\z}_i$.
Note, however, the structure of $\hat{\z}_i$'s is not taken into account if we use Eq.~\eqref{nearest}. Looking at the t-SNE embedding  visualization 
of $\hat{\z}_i$'s and $\z'_m$'s in Figure~\ref{fig:tsne} (b) (details are explained later), it can be seen that nearest neighbor will not provide an optimal label assignment. 

\subsubsection{Transductive Learning}

In the transductive attribute-aware (TAA) method,   the attributes for all test images (i.e., unseen) are first predicted to form $\hat{Z}=[\hat{\z}_1,...,\hat{\z}_L]$. Next, a graph is formed on $[Z',\hat{Z}]$, where the labels for $Z'$ are known and the task is to infer the labels of $\hat{Z}$. Intuitively, we want the data points that are close together to have similar labels. This problem can be formulated as a graph-based semi-supervised label propagation.

We follow the work of Zhou et al.~\cite{zhou2003learning} and spread the labels of $Z'$ to $\hat{Z}$. We form a graph $\mathcal{G}(\mathcal{V},\mathcal{E})$ where the set of nodes $\mathcal{V}=\{\v\}_1^{M+L}=[\z_1',...,\z_M',\hat{\z}_1,...,\hat{\z}_L]$, and $\mathcal{E}$ is the set of edges whose weights reflect the similarities between the attributes. Note that the first $M$ nodes are labeled and our task is to use  these labels to predict the labels of the rest of the nodes. We use a Gaussian kernel to measure the edge weights between the connected nodes, $W_{mn}=exp\{-\|\v_m-\v_n\|^2/2\sigma^2\}$, where $\sigma$ is the kernel parameter and $W_{ii}=0$. To construct the graph $\mathcal{G}$ one can utilize efficient k-NN graph construction methods, where the assumption is that  a neighbor of a neighbor is more likely to be a neighbor. Let $F\in\mathbb{R}^{M\times (M+L)}$ corresponds to a classification of the nodes, where $F_{mn}$ is the probability of $\v_n$ belonging to the $m$'th class. Let $Y\in \mathbb{R}^{M\times (M+L)}=[\bm{ I }_{M\times M},\bm{ 0}_{M\times L}]$ represent the initial labels, where $\bm{ I }$ denotes an identity matrix and $\bm{ 0 }$ denotes a zeros matrix. From a Graph-Signal Processing  point of view, $F$ is a signal defined on the graph $\mathcal{G}$, and one requires this signal to be smooth. Zhou et al.~\cite{zhou2003learning} proposed   to obtain a smooth signal on graph $\mathcal{G}$ that fits the initial known labels, 
\begin{equation}
\begin{split}
\operatorname{argmin}_{F}  \biggl \{\frac{1}{2}\biggl(\sum_{m,n} W_{mn}\|\frac{F_m}{\sqrt{D_{mm}}}-\frac{F_n}{\sqrt{D_{nn}}}\|^2+ \mu\sum_m\|F_m-Y_m\|^2\biggr)\biggr\}\enspace,
\label{eq:zhou}
\end{split}
\end{equation}
where $D\in\mathbb{R}^{(M+L)\times(M+L)}$ is the diagonal degree matrix of graph $\mathcal{G}$, $D_{mm}=\sum_n W_{mn}$, and $\mu$ is the fitness regularization. Note that the first term in Eq. \eqref{eq:zhou} enforces the smoothness of signal $F$ and the second term enforces the fitness of $F$ to the initial labels. Then Eq. \eqref{eq:zhou} would have the following solution:
\begin{equation}
\begin{split}
F=\frac{\mu}{1+\mu}\biggl(\mathtt{I}-\frac{1}{1+\mu}(D^{-\frac{1}{2}}WD^{-\frac{1}{2}})\biggr)^{-1}Y\enspace.
\end{split}
\end{equation}
Algorithm~\ref{algo2} summarizes the zero-shot label prediction procedure.

\section{Theoretical Discussion}
\label{sec:analysis}
In this  section, we   establish PAC-learnability of the proposed algorithm.   We provide  a PAC  style generalization error  bound for  the proposed ZSL algorithm. The goal is to establish conditions under which, our ZSL algorithm can identify instances from unseen classes. We use the framework developed by Palatucci et. al.~\cite{palatucci2009zero}, to derive this bound. The core idea is that if we are able to recover the semantic attributes of a given image with high accuracy, then the correct label can be recovered with high probability as well. Note that three probability events are involved in the probability event of predicting an unseen class label correctly, denoted by $P_t$:

1. Given a certain confidence parameter $\delta$ and error parameter $\epsilon$, a dictionary can be learned with $M_{\epsilon, \delta}$ samples. We denote this event by $\mathcal{D}_{\epsilon}$. Hence $P(\mathcal{D}_{\epsilon})=1-\delta$ and $\mathbb{E}(\|\bm{x}-D\bm{a}\|_2^2)\le\epsilon$,  where $\mathbb{E}(\cdot)$ denotes statistical expectation.

2. Given the event $\mathcal{D}_{\epsilon}$ (learned dictionaries), the  semantic attribute   can be estimated with high probability. We denote this event by $\mathcal{S}_{\epsilon}|\mathcal{D}_{\epsilon}$.

3. Given the event $\mathcal{S}_{\epsilon}|\mathcal{D}_{\epsilon}$, the true label can be predicted. We denote this event by $\mathcal{T}|\mathcal{S}_{\epsilon}$ and so $P(\mathcal{T}|\mathcal{S}_{\epsilon})=1-\zeta$.

Therefore,   the  event $P_t$  can be expressed as the following   probability decoupling by multiplying the above probabilities:
\begin{equation}
\begin{split}
P_t=P(\mathcal{D}_{\epsilon})P(\mathcal{S}_{\epsilon}|\mathcal{D}_{\epsilon})P(\mathcal{T}|\mathcal{S}_{\epsilon})\enspace.
\end{split}
\label{eq:probe1}
\end{equation}    
Our goal is   given the desired values for confidence parameters $\zeta$ and $\delta$ for the two ZSL stages, i.e., $P(\mathcal{D}_{\epsilon})=1-\delta$ and $P(\mathcal{T}|\mathcal{S}_{\epsilon})=1-\zeta$, we compute the necessary $\epsilon$  for that level of prediction confidence as well as $P(\mathcal{S}_{\epsilon}|\mathcal{D}_{\epsilon})$. We also need to compute the number of required training samples to secure the desired errors. Given $P(\mathcal{T}|\mathcal{S}_{\epsilon})=1-\zeta$, we   compute $\epsilon$     and  the conditional probability   $P(\mathcal{S}_{\epsilon}|\mathcal{D}_{\epsilon})$. 

To establish the error bound,  we need to compute the maximum error in predicting the semantic attributes of a given image, for which we still can predict the correct label with high probability. Intuitively, this error depends on geometry of $\mathcal{A}$ and probability distribution of semantic attributes of the classes in this space, $\mathcal{P}$. For example, if semantic attributes of two classes are very close, then error tolerance for those classes will be less than two classes with distant attributes. To model this intuition, we focus our analysis  on nearest neighbor label recovery. Let $\hat{\bm{z}}$ denote the predicted attribute for a given image by our algorithm. Let $d(\hat{\bm{z}},\bm{z}'):\mathbb{R}^{q}\times\mathbb{R}^{q}\rightarrow\mathbb{R}$ denote the distance between this point and another point in the semantic space. We denote the distribution function for this distance as $R_{\hat{\bm{z}}}(t)=P(d(\hat{\bm{z}},\bm{z}')\le t)$. Let $T_{\hat{\bm{z}}}$ denote    the distance to the nearest neighbor of $\hat{\bm{z}}$ and  
 $W_{\hat{\bm{z}}}(t)=P(T_{\hat{\bm{z}}}\le t)$ denotes its probability distribution. The latter distribution has been computed by Ciaccia and Patella~\cite{ciaccia2000pac} as:
\begin{equation}
\begin{split}
W_{\hat{\bm{z}}}(t)=1-\bigl(1-R_{\hat{\bm{z}}}(t)\bigl)^n\enspace,
\end{split}
\label{eq:distNN}
\end{equation}    
where $n$ is the number of points drawn from the distribution $\mathcal{P}$. Note that the function $R_{\hat{\bm{z}}}(t)$ is an empirical distribution which depends on the distribution of semantic feature space, $\mathcal{P}$, and basically is the fraction of sampled points from $\mathcal{P}$ that are less than some distance $t$ away from $\hat{\bm{z}}$. 

Following the general PAC learning framework,   given a desired probability (confidence) $\zeta$, we want the distance $T_{\hat{\bm{z}}}$ to be less than the distance of the predicted attribute $\hat{\bm{z}}$ from the true semantic description of the true class that it belongs to, i.e.,   or $W_{\hat{\bm{z}}}(\tau_{\hat{\bm{z}}})\le \zeta$. Now note that since $W_{\hat{\bm{z}}}(\cdot)$ is a cumulative distribution (never decreasing),  $W_{\hat{\bm{z}}}^{-1}(\cdot)$ is well-defined as $W_{\hat{\bm{z}}}^{-1}(\zeta)=\text{argmax}_{\tau_{\hat{\bm{z}}}}  [W_{\hat{\bm{z}}}(\tau_{\hat{\bm{z}}})\le \zeta ]$. If $\tau_{\hat{\bm{z}}}\le W_{\hat{\bm{z}}}^{-1}(\zeta)$, then the correct label can be recovered with  probability of $1-\zeta$. Hence, prior to label prediction (which itself is done for a given  confidence parameter $\delta$), the semantic attributes must be predicted with the true error at most $\epsilon_{max} = W_{\hat{\bm{z}}}^{-1}(\zeta)$ and we need to ensure that semantic attribute prediction achieves this error bound, that is $\mathbb{E}_{\z}\bigl(\|\z-D_z\a^*\|_2^2\bigr)\le W_{\hat{\bm{z}}}^{-1}(\zeta)$. To ensure this to happen, we rely on the following theorem on PAC-learnability of the dictionary learning \eqref{eq:mainDx} derived by Gribonval et. al~\cite{gribonval2015sample}:

\textbf{Theorem~1~\cite{gribonval2015sample}}:
Consider dictionary learning problem in \eqref{eq:mainDx}, and the confidence parameter $\delta$ ($P(\mathcal{D}_{\epsilon})=1-\delta$) and the error parameter $\epsilon_{max} = W_{\hat{\bm{z}}}^{-1}(\zeta)$ in standard PAC-learning setting. Then the number of required samples to learn the dictionary $M_{W_{\hat{\bm{z}}}^{-1},\delta}$ satisfies the following relation:
\begin{equation}
\begin{split}
&W_{\hat{\bm{z}}}^{-1}(\zeta) \ge 3\sqrt{\frac{\beta\log(M_{W_{\hat{\bm{z}}}^{-1},\delta})}{M_{W_{\hat{\bm{z}}}^{-1},\delta}}}+\sqrt{\frac{\beta+\log(2/\delta)/8}{M_{W_{\hat{\bm{z}}}^{-1},\delta}}}\\
&\beta=\frac{pr}{8}\text{max}\{1,\log(6\sqrt{8}L)\}\enspace,
\end{split}
\label{eq:distNN1}
\end{equation}    
where $L$ is a contestant that depends on the loss function which measures the data fidelity. Given all parameters, Eq. \eqref{eq:distNN} can be solved for $M_{W_{\hat{\bm{z}}}^{-1},\delta}$.

So, according to Theorem~1  if we use at least $M_{W_{\hat{\bm{z}}}^{-1},\delta}$ sample images to learn the coupled dictionaries, we can achieve the required error rate $\epsilon_{max} = W_{\hat{\bm{z}}}^{-1}(\zeta)$. Now we need to determine what is the probability of recovering the true label   in ZSL regime  or $P(\mathcal{S}_{\epsilon}|\mathcal{D}_{\epsilon})$. 
Note that the core step for predicting the semantic attributes in our scheme is to compute the joint sparse representation for an unseen image.   Also note that  Eq.~\ref{eq:mainDx} can  be interpreted as result of  a maximum a posteriori (MAP) inference. This means that from a probabilistic perspective, $\balpha$'s are drawn from a Laplacian distribution  and the dictionary $D$ is  a  Gaussian matrix with elements drawn i.i.d: $d_{ij} \sim  \mathcal{N}(\bm{0}, \epsilon)$. This means that given a drawn dataset, we learn MAP estimate of the Gaussian matrix $[D_x,D_z]^\top$ and then use the Gaussian matrix $D_z$ to estimate $\a$ in ZSL regime. To compute the probability of recovering $\a$ in this setting, we   rely on the following theorem:

\textbf{Theorem~2 (Theorem 3.1 in~\cite{negahban2009unified})}: Consider the linear system $\x_i=D_x\a_i+\bm{n}_i$ with a sparse solution, i.e., $\|\a_i\|_0=k$, where $D_x\in\mathbb{R}^{p\times r}$ is a random Gaussian matrix  and $\|\bm{n}_i \|_2\le\epsilon)$. Then the unique solution of this system can be recovered by solving eq.~\eqref{eq:attrAgn} with probability of $(1-e^{p\xi})$ as far as $k\le c'p\log(\frac{r}{p})$, where $c'$ and $\xi$ are two constant parameters.
 
Theorem~2 suggests that we can use eq.~\eqref{eq:attrAgn} to recover the sparse representation and subsequently unseen attributes with high probability $P(\mathcal{S}_{\epsilon}|\mathcal{D}_{\epsilon})= (1-e^{p\xi})$. This theorem also suggests that for our approach to work, existence of a good sparsifying dictionary as well as rich  attribute data is essential.   Therefore, given desired error parameters $1-\zeta$ and $1-\delta$  for the two  stages of ZSL algorithm and an error parameter $\epsilon$, the probability event of predicting an unseen class label correctly can be computed as:
\begin{equation}
\begin{split}
P_t= (1-\delta)\bigl(1-e^{p\xi}\bigr)(1-\zeta)\enspace,
\end{split}
\label{eq:finalprobe1}
\end{equation}  
which concludes our proof on PAC-learnability of the algorithm.

 \subsection{Computational Complexity} 
 A major criterion for choosing a particular approach from a set of competitors for solving a specific problem   is   computational complexity. Since in ZSL, model training is supposedly performed once, training computational complexity is a secondary criterion. As a result, it is more important to analyze the testing computational complexity. As explained, during testing, our baseline approach first solves for the joint sparse vector by solving the  LASSO problem \eqref{eq:attrAgn} and then the label is predicted from the recovered attribute using nearest neighbor. Given that $D_x\in\mathbb{R}^{p\times r}$   and $r\ge p$, the  computational complexity  for solving the  LASSO problem would be $O(r^3)$ \eqref{eq:attrAgn}~\cite{efron2004least}. Upon sparse vector recovery, we can estimate the attribute by the matrix multiplication $\hat{z}_i=D_z\balpha_i$ which has the computational complexity $O(qr)$. Finally. the nearest neighbor computational complexity would be $O(qM)$ and hence the overall computational complexity for recovering the label for an unseen class would be $O(r^3+qr+qM)$.

 To judge the efficiency of our algorithm, we can compare it against a heuristic deep learning method. For simplicity, we relax the problem and assume that due to existence of enough labeled data from the seen classes, we can train a deep networks $\mathcal{G}(\cdot):\mathbb{R}^{p}\rightarrow \mathbb{R}^{q}$ that maps a given $\bm{x}$ to the corresponding attribute $\bm{z}$ at its output. As a result, the computational complexity of using a deep net for ZSL during testing would be computational complexity of forward pass. If we denote the number of nodes in the network by $m_1,\ldots, m_n$, then computational complexity of forward pass would be  $O(pqm_1\ldots m_n)$ and assuming that, nearest neighbor is used at the output, then overall complexity would be $O(pqm_1\ldots m_n+qM)$. It is quite clear that this computational complexity can grow fast when the network is deep which is typical of the current practical network with several layers with many nodes. This demonstrates that using coupled dictionary learning is more practical if computational complexity  is a major concern.

\section{Experiments}
\label{sec:results}
We carried out experiments on four benchmark ZSL datasets and empirically evaluated the resulting performances against  existing ZSL  algorithms.

{\bf Datasets:} We conducted our experiments on four benchmark datasets namely: the Animals with Attributes   (AwA1)~\cite{lampertattribute}, (AwA2),~\cite{xian2017zero} the SUN attribute~\cite{patterson2012sun}, and the Caltech-UCSD-Birds 200-2011 (CUB)  bird~\cite{wah2011caltech} datasets. 

The AwA1 dataset is a coarse-grained dataset containing 30475 images of 50 types of animals with 85 corresponding attributes for these classes. Semantic attributes for this dataset are obtained via human annotations. The images for the AWA1 dataset are not publicly available; therefore we use the publicly available features of dimension $4096$, extracted from a VGG19 convolutional neural network, which was pretrained on the ImageNet dataset. Following the conventional usage of this dataset, 40 classes are used as the source classes to learn the model and the remaining 10 classes are  used as the target (unseen) classes to test the performance of zero-shot classification.
The major disadvantage of AwA1 dataset is that only extracted features are available for this dataset. The AwA2 dataset is developed   to compensate for this weakness by providing the original images. The  AWA2  dataset  has a similar structure with  the  same  50  animal  classes and  85 attributes, but with 37322 images. Because the original images are available, one can use alternative deepnet structures for feature extraction.

The SUN dataset is a fine-grained dataset and contains 717 classes of different scene categories with 20 images per category (14340 images total).  Each image is annotated with 102 attributes that describe the corresponding scene. There are two general approaches to split this dataset into training and testing sets. Following~\cite{zhang2015zero}, 707 classes are used to learn the dictionaries and the remaining 10 classes are used for testing. Following the second approach~\cite{lampertattribute}, we used 645 classes   to learn the dictionaries and   72 classes are used for testing. Both splits are informative because together help analyzing the effect of the training set size on  the performance. 

The CUB200 dataset is a fine-grained dataset containing 200 classes of different species of birds with 11788 images with 312 attributes and boundary segmentation for each image. The attributes are obtained via human annotation. The dataset is divided into four almost equal folds, where three folds are used to learn the model and the fourth fold is used for testing.

 For each dataset, except for AwA1 (where images are not available), we use features extracted by the final layer prior to classification of VGG19~\cite{simonyan2014very}, Inception~\cite{szegedy2015going}, ResNet~\cite{he2016deep}, and DenseNet~\cite{huang2017densely}. For AwA1, AwA2, and CUB200-2011 the networks were trained on ImageNet~\cite{krizhevsky2012imagenet}. For SUN, they were trained on Places~\cite{zhou2014learning}. 

We used   flat hit@K classification accuracy, to measure the performance. This means that a test image is said to be classified correctly  if it is classified among the top $K$ predicted labels. We report hit@1   rate to measure ZSL image classification performance and hit@3 and hit@5 for image retrieval performance.

\begin{figure*}[t]
\centering
\includegraphics[width=\linewidth]{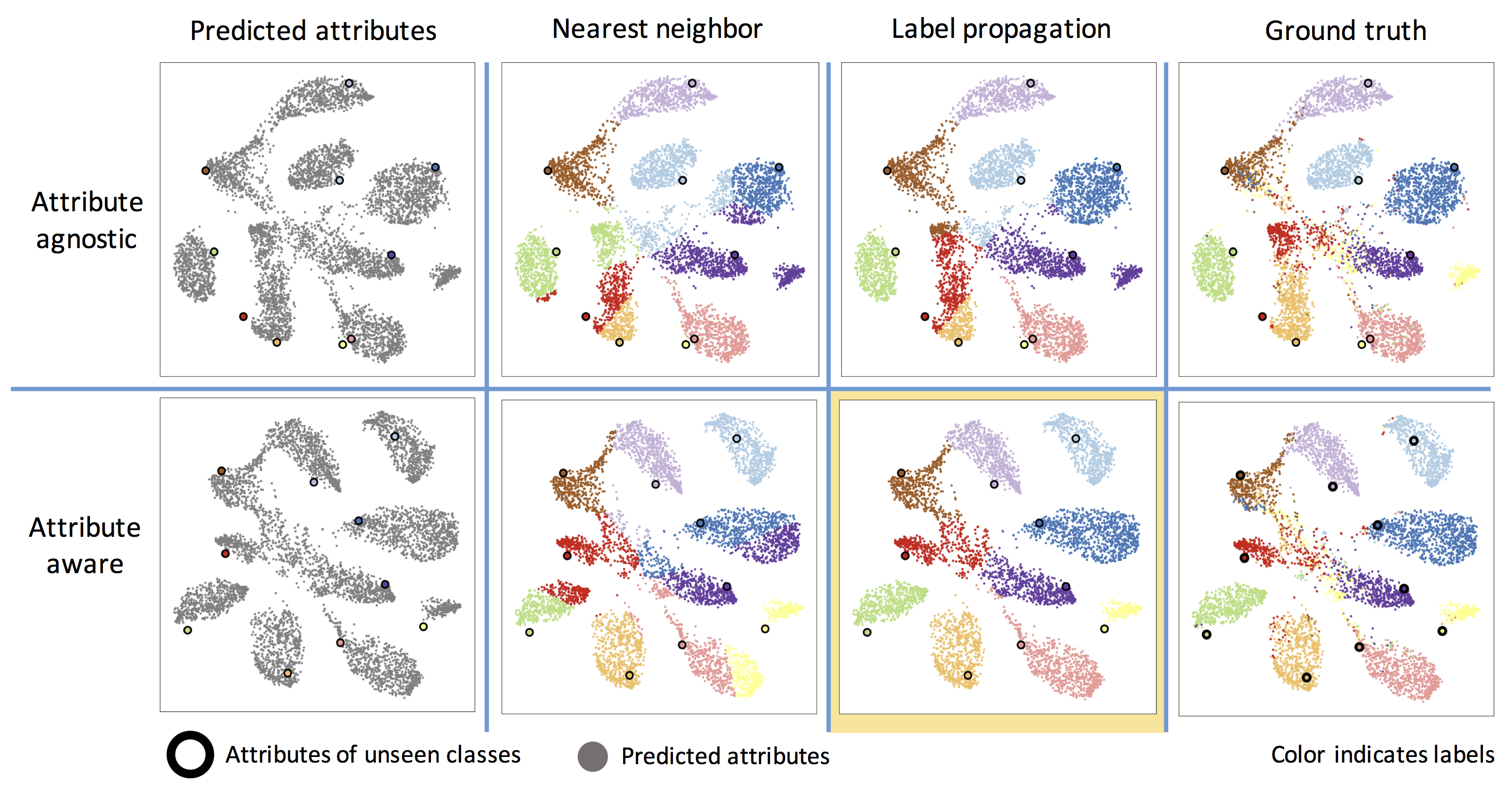}
\caption{Attributes predicted from the input visual features for the unseen classes of images for AWA1 dataset using our attribute-agnostic and attribute-aware formulations respectively in top and bottom rows. The nearest neighbor and label propagation assignment of the labels together with the ground truth labels are visualized. It can be seen that the attribute-aware formulation together with the label propagation scheme overcomes the hubness and domain-shift problems, enclosed in yellow margins. Best seen in color. }
\label{fig:tsne}
\end{figure*}

{\bf Results:} Each experiment is performed ten times and the mean is reported in Table \ref{tab:table0}. For the sake of an ablative study, we have included results for the AAg formulation using nearest neighbor, the AAw using nearest neighbor, and AAw using the transductive approach, denoted as transductive attribute-aware (TAAw) formulation. In other words, we can study the effect of the absence of each of the entropy regularization and the transductive prediction  on the performance to demonstrate their positive effect. As   can be seen, while the AAw formulation significantly improves the AAg formulation,  adding the transductive approach (i.e., label propagation on predicted attributes) to the AAw formulation further boosts the classification accuracy, as also shown in Figure \ref{fig:tsne}.   These results also support the logic behind our approach that: 1)   the attribute aware optimization always boosts the performance, and 2) the transductive prediction of labels leads to a secondary boost in performance of our method. Finally, for completeness hit@3 and hit@5 rates   measure image retrieval performance.

 \begin{table*}[t]
\centering
\scriptsize
\setlength{\tabcolsep}{1pt}
{ 
\centering
\begin{tabular}{l|cccccccccc}   
\multicolumn{1}{c}{ \diaghead{MethodDataset}{{\tiny Feature}}{{\tiny Method}}        }  & AAg \eqref{eq:attrAgn} &   AAw \eqref{eq:softass}   & TAAw & AAg(hit@3) & AAw(hit@3) & TAAw(hit@3) & AAg(hit@5) & AAw(hit@5) & TAAw(hit@5)    \\
\hline
 \multicolumn{10}{c}{AwA1 Dataset}\\
\hline
VGG19           &77.30&79.48&89.35&96.05&96.54&97.52&98.56&98.67&98.51\\
\hline
 \multicolumn{10}{c}{AwA2 Dataset}\\
\hline
VGG19           & 41.68&45.54&69.93&74.80&78.62&88.77&91.36&92.56&93.34\\
Inception       & 39.05&47.61&71.72&82.15&84.58&97.12&90.64&92.08&97.66\\
ResNet50        & 43.55&47.81&81.99&80.09&83.32&94.76&92.92&93.91&95.37\\
DenseNet161      & 40.72&43.47&78.14&77.08&80.17&98.09&94.63&95.89&98.44\\
\hline
 \multicolumn{10}{c}{CUB}\\
\hline
VGG19           &35.29&40.62&48.41&60.52&67.67&67.75&72.14&74.44&78.57\\
Inception       &35.32&40.31&49.65&51.17&55.52&63.78&67.05&71.37&75.33\\
ResNet50        &  24.81&29.79&44.19&48.22&56.52&67.03&58.93&66.69&75.60
\\
DenseNet161      &  28.91&33.55&51.03&51.51&59.57&73.13&61.06&68.25&79.63
\\
\hline
 \multicolumn{10}{c}{SUN Dataset (645/72 Split)}\\
\hline
 
VGG19           &42.36&45.69&48.40&57.50&61.48&67.50&71.94&75.76&82.01 

\\
Inception       &55.66&56.02&57.03&80.10&80.65&81.18&87.06&87.22&87.72

\\
ResNet50        & 44.60&45.49&53.09&70.13 &70.76 &75.06 &79.53&79.81&81.79
\\
DenseNet161      & 42.76&43.48&51.22&68.24&68.76&74.65&77.71&78.40&81.35
\\
\hline
 \multicolumn{10}{c}{SUN Dataset (707/10 Split)}\\
\hline
 
VGG19           &  85.50&  89.25& 91.00& 93.95& 96.50& 98.05& 97.15& 98.05&98.50
\\
Inception       &  83.30& 83.80& 84.95& 96.80& 96.80& 96.95& 98.85& 98.85& 98.80
\\
ResNet50        &  76.10&83.60&84.60&93.20&97.35&97.10&96.70&96.95&99.05
\\
DenseNet161      &  74.65&75.10&86.65&93.05&93.05&97.30&96.60&96.70&99.25
\\
\end{tabular}}
\caption{ Zero-shot classification and image retrieval results for the proposed algorithm.  }
\label{tab:table0}
\end{table*}
\normalsize

Figure \ref{fig:tsne} demonstrates the 2D t-SNE embedding  for predicted attributes and actual class attributes of the AWA1 dataset.  It can be seen  that our algorithm can cluster the dataset in the attribute space.   The actual attributes are depicted by the colored circles with black edges. The first column of Figure \ref{fig:tsne} demonstrates the attribute prediction for AAg and AAw formulations. It can be   seen that the entropy regularization in AAw formulation improves the clustering quality, decreases data overlap, and reduces the domain-shift problem. The nearest neighbor label assignment is shown in the second column, which demonstrates the domain-shift and hubness problems with NN label assignment in the attribute space. The third column of Figure \ref{fig:tsne} shows the transductive approach in which a label propagation is performed on the graph of the predicted attributes. Note that the label propagation addresses the domain-shift and hubness problem and when used with the AAw formulation improves the accuracy significantly.

 \begin{table}[t!]
 \centering
{\small
\begin{tabular}{lc|ccc}   
\multicolumn{2}{c}{Method}    & SUN & CUB & AwA1    \\
\hline
\multicolumn{2}{c|}{Romera-Paredes and Torr~\cite{romera2015embarrassingly}  }&  82.10 & - &  75.32 	\\
\multicolumn{2}{c|}{Zhang and Saligrama~\cite{zhang2015zero}$^\dagger$}	& 82.5 & 30.41 & 76.33 	\\
\multicolumn{2}{c|}{Zhang and Saligrama~\cite{zhang2016zero}$^\dagger$}      &      83.83   &   42.11  & 80.46\\
\multicolumn{2}{c|}{Bucher, Herbin, and Jurie~\cite{bucher2016improving}$^\dagger$}&84.41& 43.29 &  77.32\\
\multicolumn{2}{c|}{Xu et. al.~\cite{xu2017matrix}$^\dagger$}& 84.5 &  53.6 & 83.5   \\
\multicolumn{2}{c|}{Long et. al.~\cite{long2018zero} $^\dagger$}		& 80.5 & - & 82.12   \\
\multicolumn{2}{c|}{Lu et. al.~\cite{lu2018attribute} $^\dagger$}		& 84.67 & 44.67 & 65.89   \\
\multicolumn{2}{c|}{Ye and Guo~\cite{yezero}$^\dagger$ }		&  85.40 &  57.14 & 87.22   \\
\multicolumn{2}{c|}{Ding,  Shao,  and Fu~\cite{ding2017lowrank}$^\dagger$ }		&  86.0 &  45.2 & 82.8   \\
\multicolumn{2}{c|}{Wang and Chen~\cite{wang2017zero}$^\dagger$} &- & 42.7&79.8  \\
\multicolumn{2}{c|}{Kodirov,  Xiang,    and Gong ~\cite{kodirov2017semantic}$^\dagger$} &91.0 & 61.4 &84.7\\ 
\multicolumn{2}{c|}{Ding et. al. ~\cite{ding2018generative}$^\dagger$} &88.2 & 48.50 &84.74\\
\hline
Ours & AAg \eqref{eq:attrAgn}              &85.5    & 35.29 &  77.30  \\
Ours & AAw \eqref{eq:softass}             & 89.3  & 40.62 &   79.48  \\
Ours & Transductive AAw (TAAw)            & 91.00 & 48.41 &   89.35  \\
\end{tabular}}
\caption{ Zero-shot classification results for four benchmark datasets. All methods use VGG19 features trained on the ImageNet dataset and the original continuous (or binned) attributes provided by the datasets. Here, $\dagger$ indicates that the results are extracted directly from the corresponding paper, $\ddagger$ indicates that the results are reimplemented with VGG19 features, and $-$ indicates that the results are not reported. }
\label{tab:table1}
\end{table}

 \begin{table}[t!]
 \centering
{\small
\begin{tabular}{lc|ccc }   
\multicolumn{2}{c}{Method}    & SUN & CUB & AwA2    \\
\hline
\multicolumn{2}{c|}{Romera-Paredes and Torr~\cite{romera2015embarrassingly}$^\dagger$} &  18.7  & 44.0  & 64.5  	\\ 
\multicolumn{2}{c|}{Norouzi et. al.~\cite{norouzi2013zero}$^\dagger$}	& 51.9  &  36.2 &  63.3 	\\ 
\multicolumn{2}{c|}{Mensink et. al.~\cite{mensink2014costa}$^\dagger$}      & 47.9         &    40.8  & 61.8 \\ 
\multicolumn{2}{c|}{Akata et. al.~\cite{akata2015evaluation}$^\dagger$}&56.1 & 50.1 &  66.7\\ 
\multicolumn{2}{c|}{Lampert et. al.~\cite{lampert2014attribute}$^\dagger$}& 44.5 &  39.1 & 60.5   \\
\multicolumn{2}{c|}{Changpinyo et. al.~\cite{changpinyo2016synthesized} $^\dagger$}		& 62.7 & 54.5 & 72.9    \\ 
\multicolumn{2}{c|}{Bucher, Herbin, and Jurie~\cite{bucher2016improving}$^\dagger$ }		& -   &  43.3  &  77.3   \\
\multicolumn{2}{c|}{Xian et. al.  ~\cite{xian2016latent}$^\dagger$ }		&  - &  45.5 & 71.9   \\ 
\multicolumn{2}{c|}{Bucher et. al.~\cite{bucher2017generating}$^\dagger$} &56.4 & 60.1&55.3  \\
\multicolumn{2}{c|}{Zhang and Saligrama~\cite{zhang2015zero}$^\dagger$} &- &30.4   & 76.3\\
\multicolumn{2}{c|}{Long et al.~\cite{long2018zeroo}$^\dagger$}&-& 58.40 &  79.30\\
\hline
Ours & AAg \eqref{eq:attrAgn}              & 55.7 &35.3  &  39.1  \\
Ours & AAw \eqref{eq:softass}             & 56.0 & 40.3 &    47.6 \\
Ours & Transductive AAw (TAAw)            & 57.0 &49.7  &   71.7  \\
\end{tabular}}
\caption{ Zero-shot classification results for three benchmark datasets. All methods use Inception features trained on the ImageNet dataset and the original continuous (or binned) attributes provided by the datasets. Here  $-$ indicates that the results are not reported. }
\label{tab:table3}
\end{table}

Performance comparison results using VGG19 and GoogleNet extracted features are summarized  in Table \ref{tab:table1} and and Table \ref{tab:table3}. 
  Note that in Table \ref{tab:table3} we used AwA2 dataset in order to  extract the ResNet and GoogleNet features. As pointed out by Xian et al.~\cite{xian2017zero} the variety of used image features (e.g., various DNNs and various combinations of these features) as well as the variation of used attributes (e.g., word2vec, human annotation), and different data splits make direct comparison with the ZSL methods in the literature very challenging. In Table \ref{tab:table1}  and Table \ref{tab:table3} we provide  comparison of our JDZSL performance to the recent methods in the literature. All compared methods use the same visual features  and the same attributes (i.e., the continuous or binned) provided in the dataset to make the comparison fair. 
Table \ref{tab:table1}  and Table \ref{tab:table3} conclude the following remarks:
\begin{enumerate}
\item Comparing Table \ref{tab:table1}  with Table \ref{tab:table3} reveals that the visual features affect the ZSL performance significantly. This is a natural observation as ZSL depends on how discriminative the features are across different classes. 
\item We observe that our method achieves state-of-the-art or close to the state-of-the-art performance for both zero-shot scene and object recognition tasks. Quite importantly, while some of the other methods perform better on a specific dataset,  our algorithm leads to competitive performance on all the four benchmark datasets.
\item We observe that despite not using a deep neural network for modelling the cross-domain mapping function, we are able to achieve a competitive performance. This observation concludes that our method can potentially work better than some of the recent deep learning-based algorithms when instances from the seen classes are limited.
\item Considering progressive improvement of our results when using AAw and TAAw solutions, we conclude that secondary mechanisms to address hubness~\cite{dinu2014improving} and domain-shift~\cite{fu2015transductive} are necessary to improve ZSL algorithms.
\end{enumerate}

\section{Conclusions and Discussions}
\label{sec:conclusion}
In this paper, we developed a new zero-shot learning (ZSL) algorithm by recasting the ZSL problem as a coupled dictionary learning problem. In our formulation,  the relationship between  visual features  and semantic attributes of data points are captured via a shared sparse representation vector in the two dictionary domains.
We can use this formulation because representing signals that share some level of commonality in a union of subspaces is feasible. In the ZSL setting, since we primarily focus on classifying classes within one domain, they share a good level of commonalities.
As a result of learning the two dictionaries, the shared sparse domain acts as a shared embedding space that is used to map images to their semantic descriptions.  We established theoretical results for PAC-learnability of our method. Our analysis supports that training these two dictionaries given a sufficient number of samples is feasible. 

In addition to the baseline algorithm based on CDC, we also  demonstrated that  an entropy regularization scheme can help with  the domain-shift. We face domain-shift on ZSL because the dictionaries are trained primarily 
 based on the seen classes. As a result, a recovered sparse representation vector is biased towards the representations for seen classes.
Entropy regularization is helpful to tackle domain-shift because it biases the recover sparse vector to be close to the sparse representations of attributes of an unseen class. As a result, domain-shift challenge is mitigated.
Our results also demonstrate that a transductive approach towards assigning labels to the predicted attributes can boost the performance considerably and lead to state-of-the-art zero-shot classification by mitigating the hubness challenge.
Hubness challenge occurs as a result of curse of dimensionality which makes Euclidean distance a non-perfect measure of similarity because it is only a point-wise measure of similarity.  Our trandusctive approach considers structure of the data points to compute similarity to mitigate hubness in high dimensions. Our empirical ablative results demonstrate that both approaches are effective.
We also compared our method with the   state-of-the-art approaches in the literature and demonstrated its competitiveness on four primary ZSL benchmark datasets. An advantage of our method is that it preforms decently all the four datasets, despite the diversity between these datasets. 

Our method is not an end-to-end training  method and requires preprocessed  suitable visual and semantic features. This limits the applicability of our method in situations that suitable feature extraction methods are not accessible. Note, however, in many common domains,   pretrained models are able to generate discriminative features.   The upside of our method is that dictionary learning is less data-greedy  compared to the end-to-end methods based on deep learning both in terms of training and also during model execution. Hence, compared to the ZSL methods that use deep neural networks, dictionary learning is more effective when the number of annotated training data is small. 
An unexplored aspect for future work is extension to generalized ZSL setting. In a generalized ZSL setting, we encounter samples from both seen and unseen classes  during testing.  
As a result, domain-shift will become a more challenging obstacle because the  model is biased to recover sparse respresentations of the seen classes.

  \bibliographystyle{apa}


 

\end{document}